# ARTICLE INFORMATION

**Article title**

One-Stage-TFS: Thai One-Stage Fingerspelling Dataset for Fingerspelling Recognition Frameworks

**Authors**

Siriwiwat Lata[a], Sirawan Phiphitphatphaisit[b], Emmanuel Okafor[c], Olarik Surinta[d,*]

**Affiliations**

[a] Department of Communication Arts, Faculty of Informatics, Mahasarakham University, Mahasarakham, 44150 Thailand

[b] Department of Information System, Faculty of Business Administration and Information Technology, Rajamangala University of Technology Isan Khon Kaen Campus, Khon Kaen 40000, Thailand

[c] SDAIA-KFUPM Joint Research Center for Artificial Intelligence, King Fahd University of Petroleum and Minerals, Dharan 31261, Saudi Arabia

[d] Multi-agent Intelligent Simulation Laboratory (MISL) Research Unit, Department of Information Technology, Faculty of Informatics, Mahasarakham University, Mahasarakham, 44150 Thailand

**Corresponding author's email address**

Email address: olarik.s@msu.ac.th (O. Surinta)

**Keywords**

One-stage fingerspelling; Fingerspelling recognition; Hand detection; Hand gesture recognition; Deep learning; Computer vision

**Abstract**

The Thai One-Stage Fingerspelling (One-Stage-TFS) dataset is a comprehensive resource designed to advance research in hand gesture recognition, explicitly focusing on the recognition of Thai sign language. This dataset comprises 7,200 images capturing 15 one-stage consonant gestures performed by undergraduate students from Rajabhat Maha Sarakham University, Thailand. The contributors include both expert students from the Special Education Department with proficiency in Thai sign language and students from other departments without prior sign language experience. Images were collected between July and December 2021 using a DSLR camera, with contributors demonstrating hand gestures against both simple and complex backgrounds. The One-Stage-TFS dataset presents challenges in detecting and recognizing hand gestures, offering opportunities to develop novel end-to-end recognition frameworks. Researchers can utilize this dataset to explore deep learning methods, such as YOLO, EfficientDet, RetinaNet, and Detectron, for hand detection, followed by feature extraction and recognition using techniques like convolutional neural networks, transformers, and adaptive feature fusion networks. The dataset is accessible via the Mendeley Data repository and supports a wide range of applications in computer science, including deep learning, computer vision, and pattern recognition, thereby encouraging further innovation and exploration in these fields.

# SPECIFICATIONS TABLE

| | |
|---|---|
| **Subject** | Computer Science |
| **Specific subject area** | Fingerspelling recognition is a crucial component of hand gesture recognition frameworks. The Thai One-Stage Fingerspelling Dataset includes images and annotated files that indicate the location of the hand within the images. This dataset is designed to support the detection of hand gestures and the recognition of fingerspelling. Additionally, it is relevant to various fields, including deep learning, computer vision, pattern recognition, and computer science applications. |
| **Type of data** | Image (JPG format) and Raw (XML format) |
| **Data collection** | The Thai One-Stage Fingerspelling (One-Stage-TFS) Dataset provides fingerspelling in the Thai language, focusing exclusively on one-stage consonants. The dataset was collected between July and December 2021 by undergraduate students at Rajabhat Maha Sarakham University, Thailand. The contributors included students from the Special Education Department with experience in Thai sign language and students from other departments without experience. |
| **Data source location** | Institution: Rajabhat Maha Sarakham University<br>Province: Mahasarakham<br>Country: Thailand |
| **Data accessibility** | Repository name: Mendeley Data<br>Data identification number: 10.17632/rknd3wbz42.1<br>Direct URL to data: https://data.mendeley.com/datasets/rknd3wbz42/1 |
| **Related research article** | S. Lata, O. Surinta, An end-to-end Thai fingerspelling recognition framework with deep convolutional neural networks, ICIC Express Letters 16 (2022) 529–536. https://doi.org/10.24507/icicel.16.05.5529 |

# VALUE OF THE DATA

- In 2021, the Thai fingerspelling images were collected using a DSLR camera. This dataset focuses exclusively on one-stage fingerspelling of Thai consonants, comprising 7,200 images representing 15 Thai signs. The images have a resolution of 1280x720 pixels. The contributors include undergraduate students from the Special Education Department with experience in Thai sign language and students from other departments without experience.

- The Thai One-Stage Fingerspelling (One-Stage-TFS) dataset provides a resource for two research tasks: locating the hand within the image against complex backgrounds and recognizing the hand configuration used for fingerspelling. Additionally, the dataset does not specify whether the hands in the images are left or right, adding another layer of complexity.
- The One-Stage-TFS dataset is designed to support research in computer science and related fields. It offers a unique opportunity to develop novel end-to-end frameworks for fingerspelling recognition, encompassing both detection and recognition. Researchers can use this dataset to create frameworks that improve accuracy and reduce computational time for detecting and classifying Thai one-stage fingerspelling.

# BACKGROUND

Deafness and hearing loss are global issues that affect approximately 34 million people worldwide, according to the World Health Organization (WHO), and they necessitate effective communication methods such as sign language [1]. In Thailand, Rajabhat Maha Sarakham University has established a curriculum within the Special Education Department that allows students to obtain a bachelor's degree in sign language. Note that Mahasarakham Province is located in northeast Thailand, as illustrated in Fig. 1. The research team had the opportunity to collaborate with an experienced teacher in Thai sign language, who provided consultation as we collected Thai fingerspelling data. We focused on one-stage fingerspelling, encompassing 15 Thai consonants, each represented by a single static sign. Fourteen undergraduate students from the Special Education Department, all experienced in Thai sign language, assisted in gathering the data. Additionally, we enlisted the help of 50 students with no prior experience in sign language, who were instructed to replicate hand signs based on provided examples. The resulting Thai One-Stage Fingerspelling (One-Stage-TFS) dataset includes 7,200 images representing 15 Thai consonants.

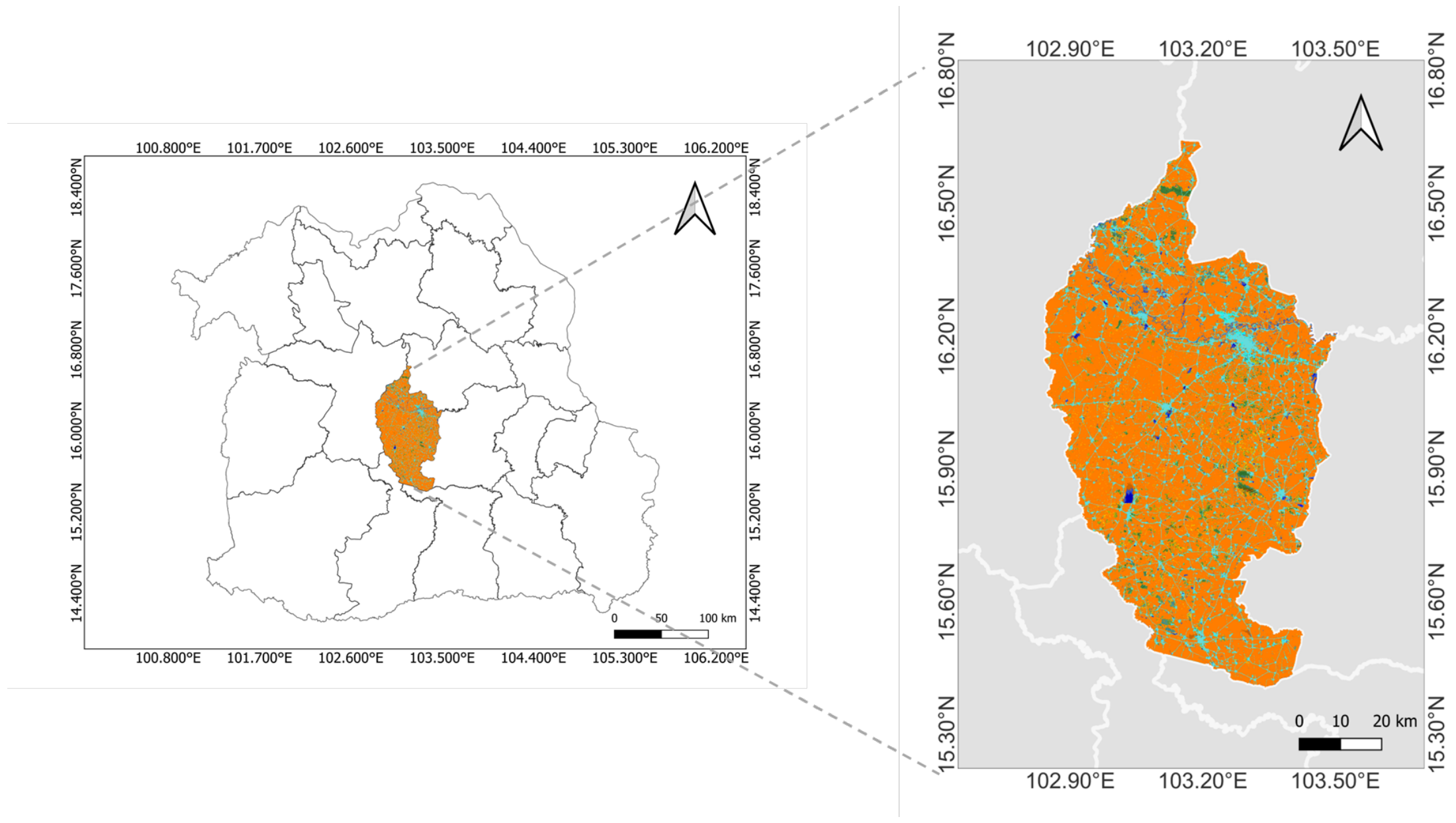

(a) Northeast Thailand (b) Mahasarakham Province

**Fig. 1.** Map showing Mahasarakham Province located in northeast Thailand.

# DATA DESCRIPTION

In early 2021, the research team consulted with a teacher from the Special Education Department at Rajabhat Maha Sarakham University, Thailand. Based on the consultation, we decided to focus on collecting Thai fingerspelling data, specifically one-stage fingerspelling, which includes 15 Thai consonants, each represented by a single static sign. Data collection took place from July and December 2021, we commenced the data collection process, using a DSLR camera to capture the hand gestures of contributors. These contributors consisted of experienced students from the Special Education Department who were proficient in Thai sign language and students from other departments with no prior experience in sign language. The captured images included upper-body shots, which could feature either one or both hands.

In the initial phase, the research team photographed the 15 Thai consonant fingerspelling signs performed by 14 proficient in Thai sign language. These images were then presented as examples to 40 students, who subsequently replicated the Thai consonant fingerspelling signs. In the second phase, we photographed these 40 students as they performed the signs. The One-Stage-TFS dataset ultimately consists of 7,200 images of Thai fingerspelling. The dataset includes images of the contributors taken against simple and complex backgrounds, as illustrated in Fig. 2. Moreover, the position of the hand varies across different locations.

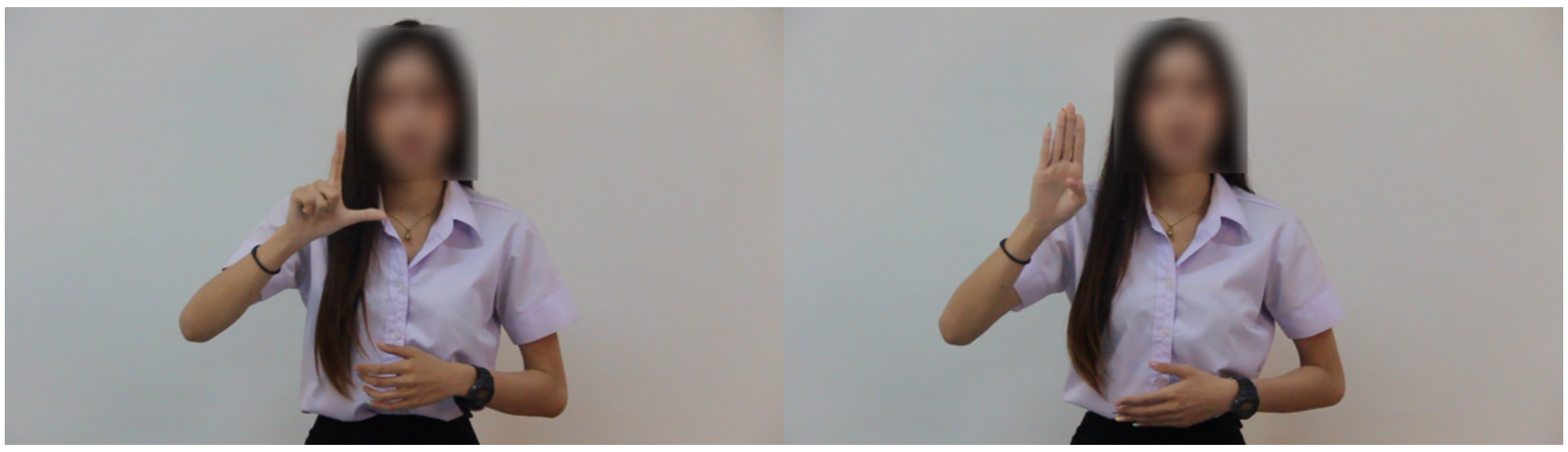

(a)

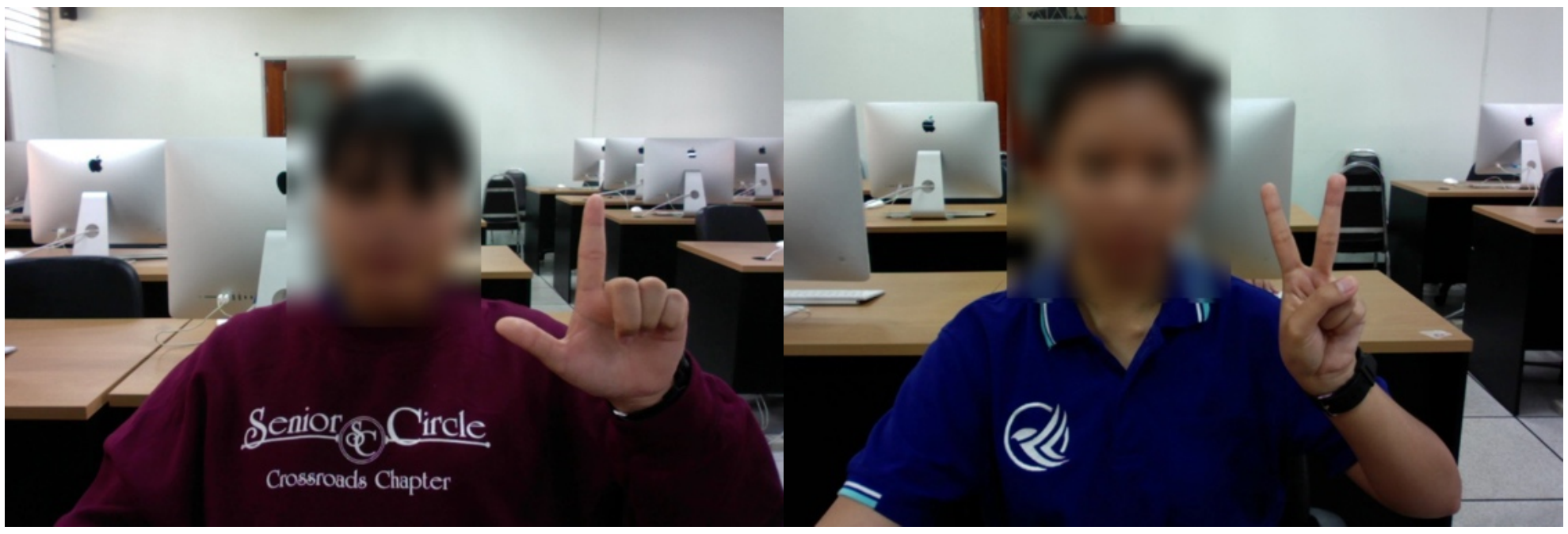

(b)

**Fig. 2.** Challenges in hand detection within the One-Stage-TFS dataset: images of contributors taken against (a) a simple background and (b) a complex background.

# EXPERIMENTAL DESIGN, MATERIALS AND METHODS

### Materials

The objective of collecting the One-Stage-TFS dataset was to capture Thai sign language, specifically focusing on the one-stage fingerspelling of 15 Thai consonants: Bor Bai Mai (/b/), Dor Dek (/d/), For Fun, (/f/), Hor Heep (/h/), Kor Kai (/k/), Lor Ling (/l/), Mor Mah (/m/), Nor Noo (/n/), Or Ang (/o/), Por Phan (/$p^h$/), Rao Ruea (/r/), Sor Suea (/s/), Tor Tao (/t/), Wor Waen (/w/), and Yor Yak (/y/). The contributors to this research included two groups of students: (1) 14 expert students from the Special Education Department at Rajabhat Maha Sarakham University, Thailand, and (2) 50 students without any experience in sign language. The dataset is publicly accessible via the Mendeley Data repository (https://data.mendeley.com/datasets/rknd3wbz42/1).

In the first process, 14 expert students in Thai sign language demonstrated 15 one-stage consonant gestures. Images were then captured of the upper body of each contributor, which could include one or both hands, with each one-stage consonant represented by a gesture from a single hand. Example images from the expert students in Thai sign language are shown in Fig. 3.

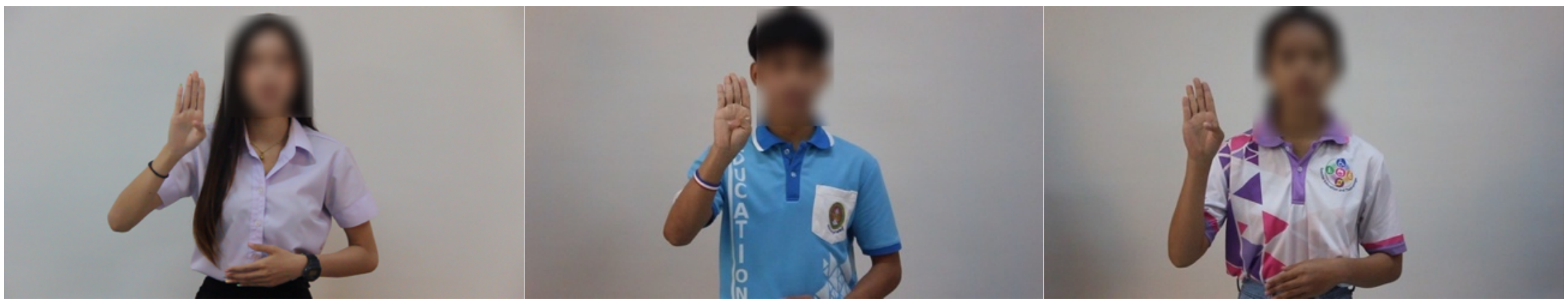

(a) Bor Bai Mai, /b/

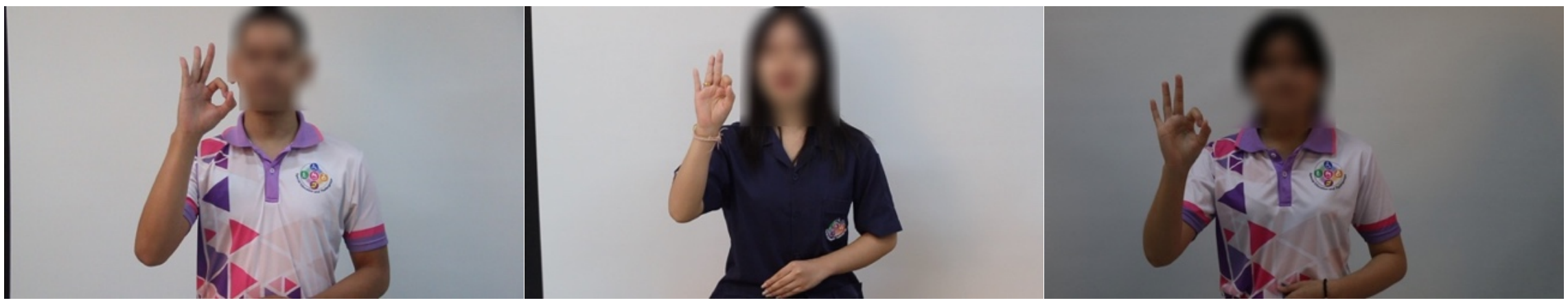

(b) For Fun, /f/

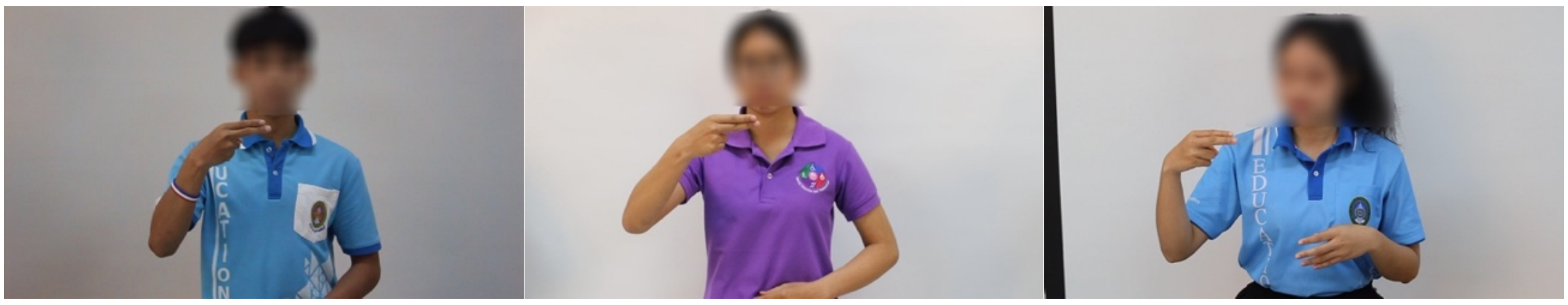

(c) Hor Heep, /h/

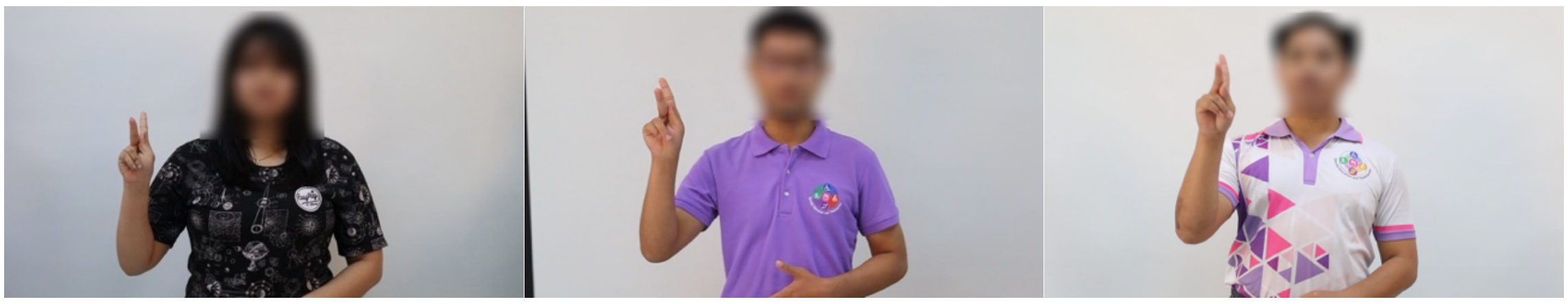

(d) Kor Kai, /k/

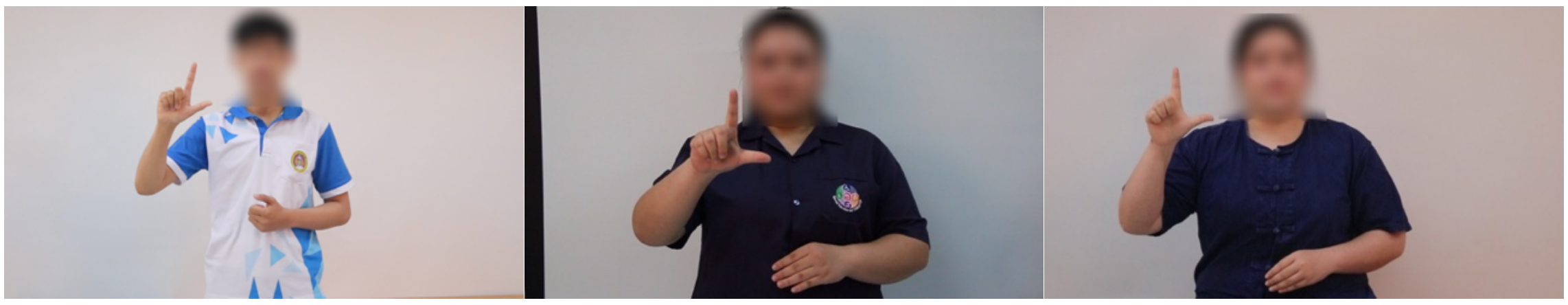

(e) Lor Ling, /l/

**Fig. 3.** Example of one-stage consonant images from the One-Stage-TFS dataset, contributed by students from the Special Education Department with experience in Thai sign language.

In the second process, we presented the images from the first process to 50 other students who had no background in sign language and instructed them to replicate the hand signs based on the provided examples, as shown in Fig. 4.

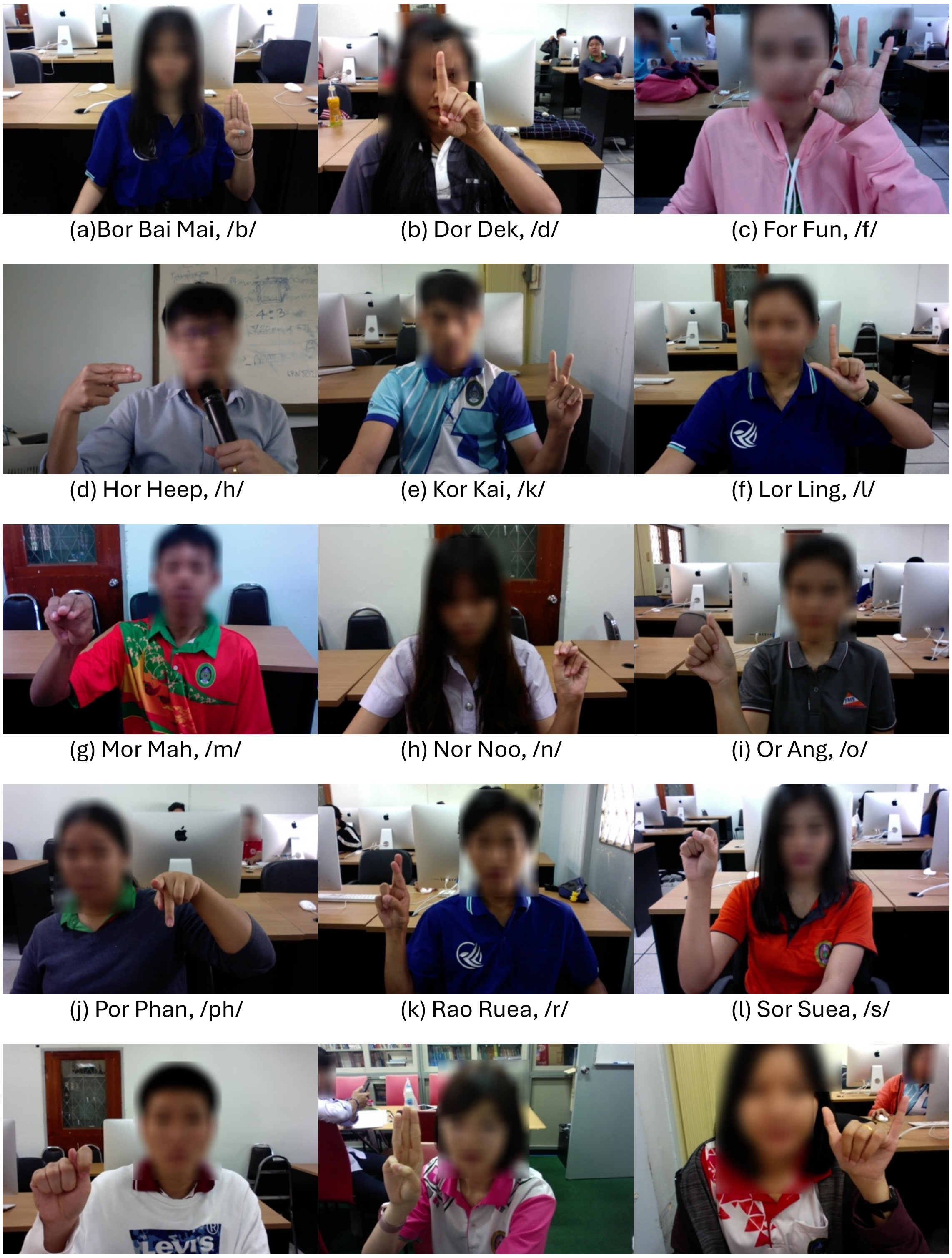

(a)Bor Bai Mai, /b/ (b) Dor Dek, /d/ (c) For Fun, /f/

(d) Hor Heep, /h/ (e) Kor Kai, /k/ (f) Lor Ling, /l/

(g) Mor Mah, /m/ (h) Nor Noo, /n/ (i) Or Ang, /o/

(j) Por Phan, /ph/ (k) Rao Ruea, /r/ (l) Sor Suea, /s/

(m) Tor Tao, /t/ (n) Wor Waen, /w/ (o) Yor Yak, /y/

**Fig. 4.** Example of one-stage consonant images, contributed by students without experience in Thai sign language.

In the third process, we repeated the second process but randomly selected students without a sign language background. The students replicated the hand signs, and we captured the images, which we used as unseen set. Examples of the images taken during this process are shown in Fig. 5.

We addressed concerns about sensitive information by blurring the faces of individuals in the One-Stage-TFS dataset, as shown in Fig. 2 - Fig. 5.

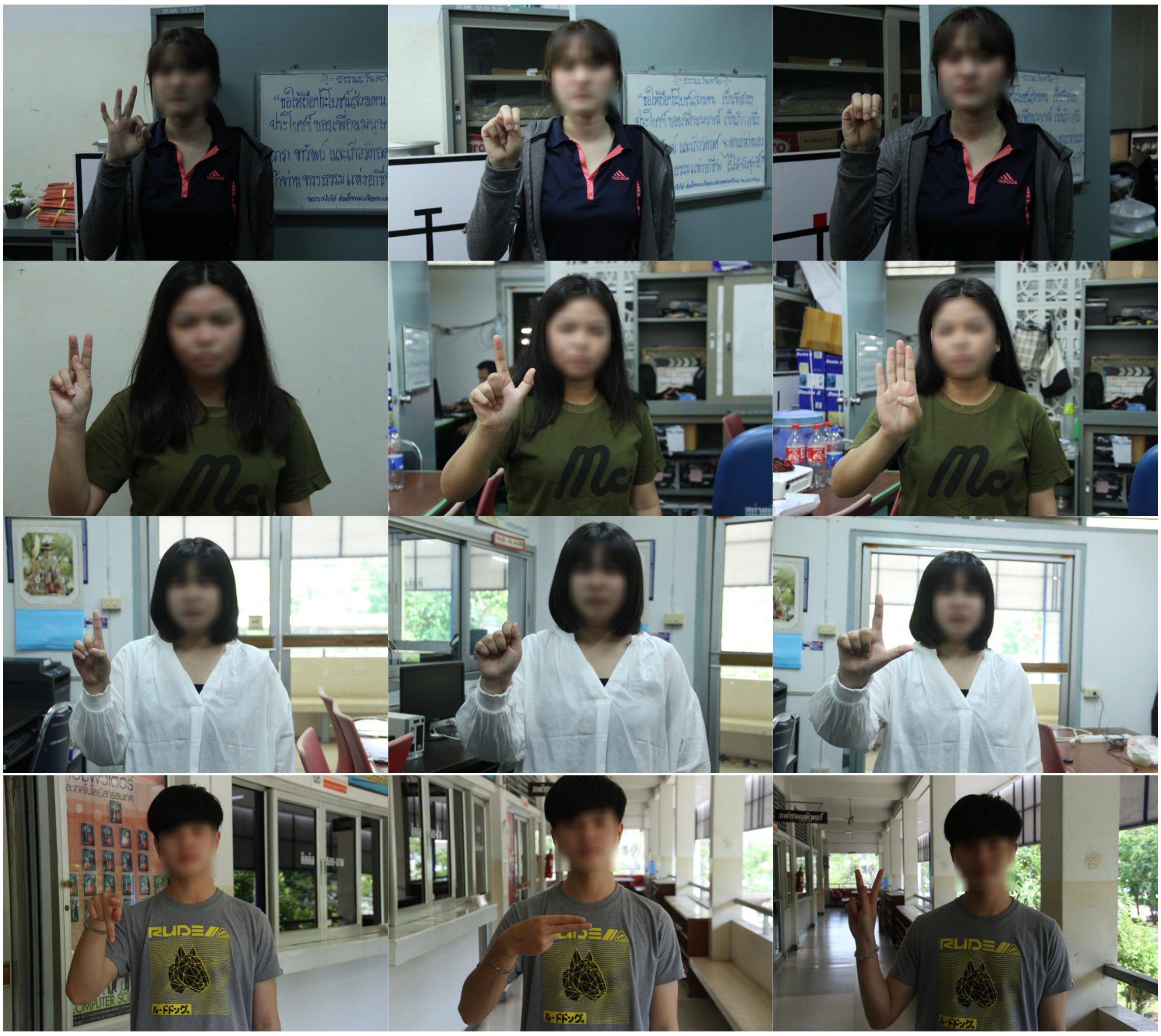

**Fig. 5.** Examples of one-stage consonant images from the unseen set.

In this dataset, ground truth labels were assigned to each hand sign image at the hand location using LabelIMG software. If researchers wish to experiment solely on Thai fingerspelling recognition, they will need to code a program to read information such as the hand sign label and hand location from the XML file. For tasks involving both detection and recognition, researchers can utilize the images and XML files during model training. An example of a ground truth file is illustrated in Fig. 6.

The file structure of the One-Stage-TFS dataset available in the Mendeley Data repository is presented in Fig. 7.

```
<annotation>
	<folder>05_BOR</folder>
	<filename>BOR_BAI_MAI_2.jpg</filename>
	<path>/Volumes/bank_cg/Banky_WS/dataset_test_v1/05_BOR/BOR_BAI_MAI_2.jpg</path>
	<source>
		<database>Unknown</database>
	</source>
	<size>
		<width>1080</width>
		<height>720</height>
		<depth>3</depth>
	</size>
	<segmented>0</segmented>
	<object>
		<name>bor</name>
		<pose>Unspecified</pose>
		<truncated>0</truncated>
		<difficult>0</difficult>
		<bndbox>
			<xmin>694</xmin>
			<ymin>241</ymin>
			<xmax>827</xmax>
			<ymax>526</ymax>
		</bndbox>
	</object>
</annotation>
```

**Fig. 6.** Example of a XML file detailing the image size, hand location, and corresponding hand sign label.

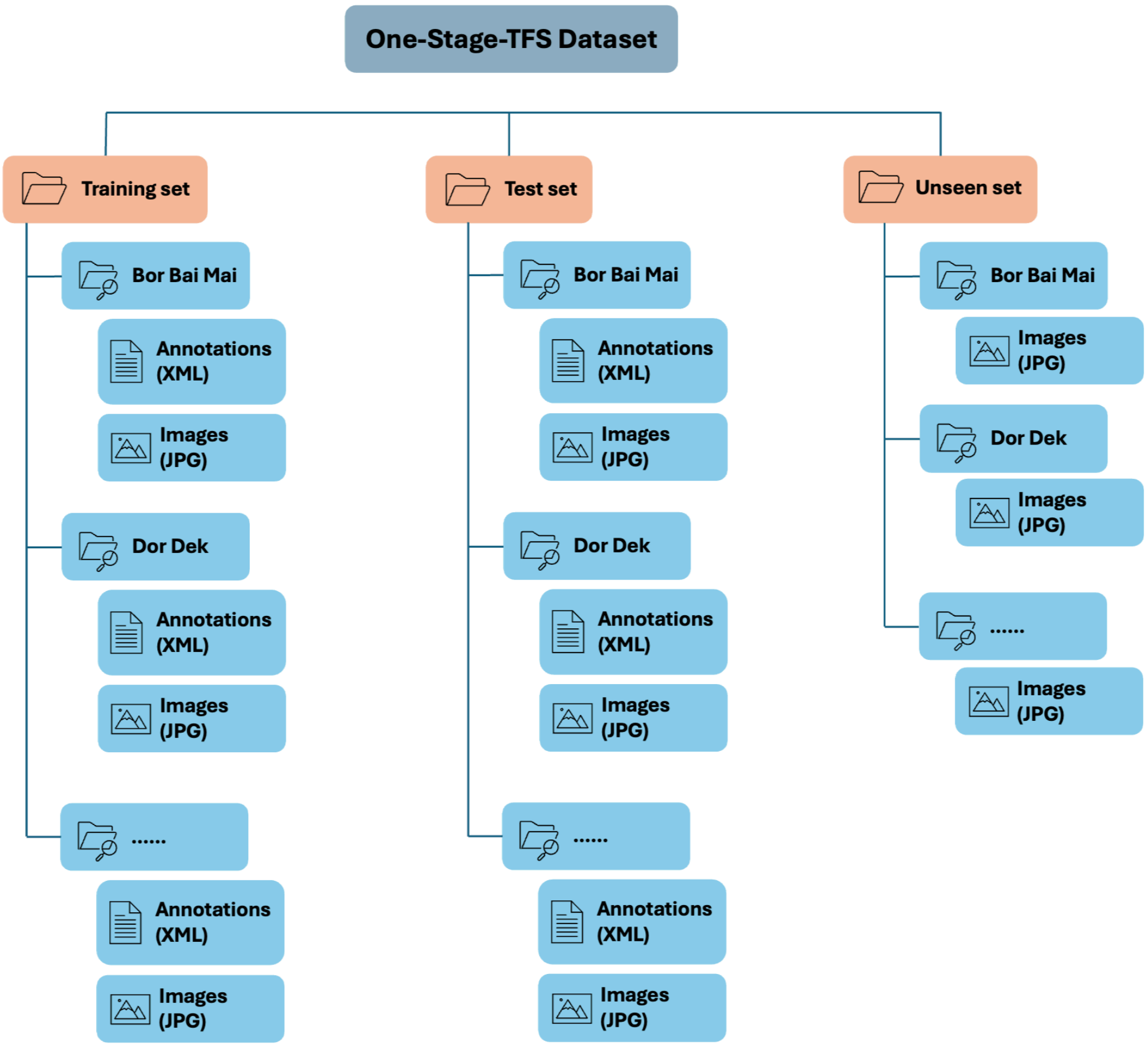

**Fig. 7.** Structure and format of files in the Mendeley Data repository.

## Experimental design

An end-to-end fingerspelling recognition framework seamlessly processes raw data, such as images, to directly predict the expected output, specifically the recognition of hand gestures in sign languages. In 2022, Lata and Surinta [2] proposed an end-to-end Thai fingerspelling recognition framework capable of detecting hands within images, regardless of the location of the hands. They employed the YOLOv3 algorithm in their experiment, which resulted in a high average precision value with an mAP of 0.9787. Although the YOLO algorithm typically handles both detection and recognition, in their approach, the label was set to "hand" rather than specifying individual fingerspelling classes. The detected hand images were subsequently processed by convolutional neural networks (CNNs) using various architectures: DenseNet121, MobileNetV2, InceptionResNetV2, NASNetMobile, and EfficientNetB2. Their experiments revealed that MobileNetV2 outperformed the other CNN models when the training set was limited, achieving an accuracy of 93.86% with only 30% of the training set. Additionally, DenseNet121 slightly outperformed MobileNetV2 when trained on 40%, 50%, and 60% of the training set. Although the recognition accuracy was below 95%, this outcome indicates the necessity for further research to develop innovative methods that exceed this performance benchmark.

In the field of computer science, the potential applications of the fingerspelling recognition are extensive, as illustrated in the framework illustrated in Fig. 8.

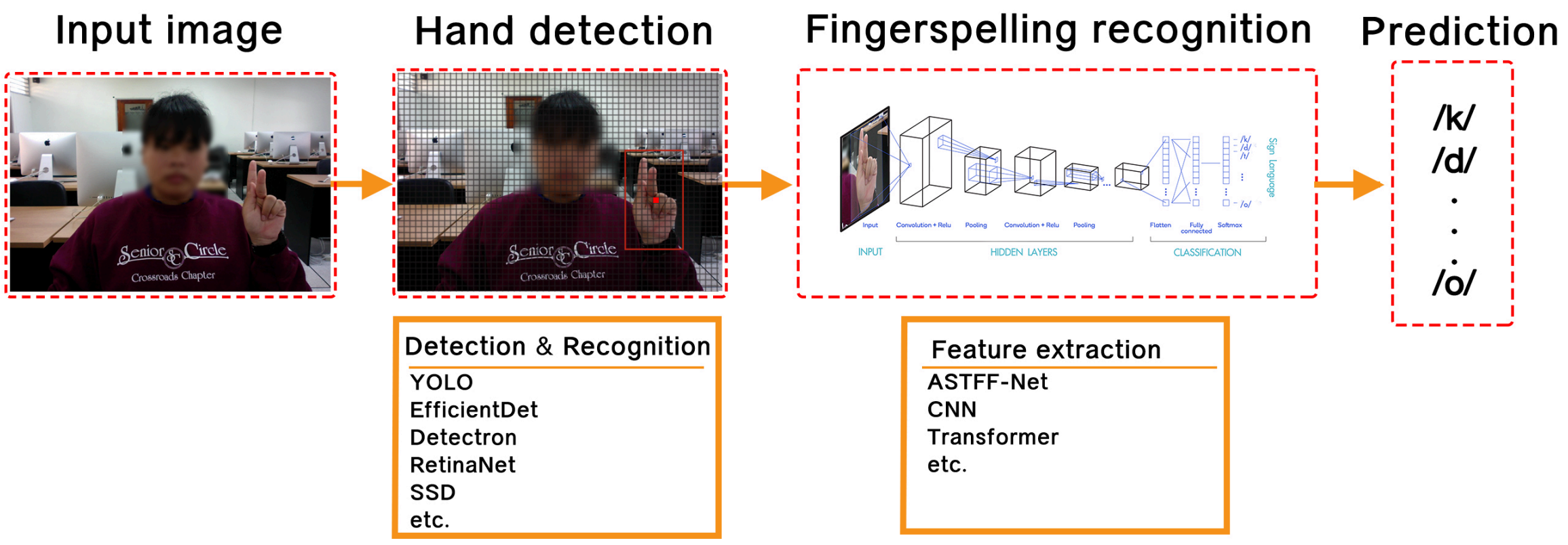

**Fig. 8.** The framework for end-to-end fingerspelling recognition.

Researchers can employ deep learning methods such as YOLO, EfficientDet, RetinaNet, Detectron [3–5] to detect and recognize hands. Examples of hand detection are shown in Fig. 9. After detection, robust features can be extracted and the objects recognized using various techniques, including CNNs, long short-term memory, (LSTM), transformers, adaptive feature fusion networks, and multi-layer adaptive spatial-temporal feature fusion networks (ASTFF-Net) [6–9]. This dataset opens new avenues for research and development in computer science, encouraging further exploration of its possibilities.

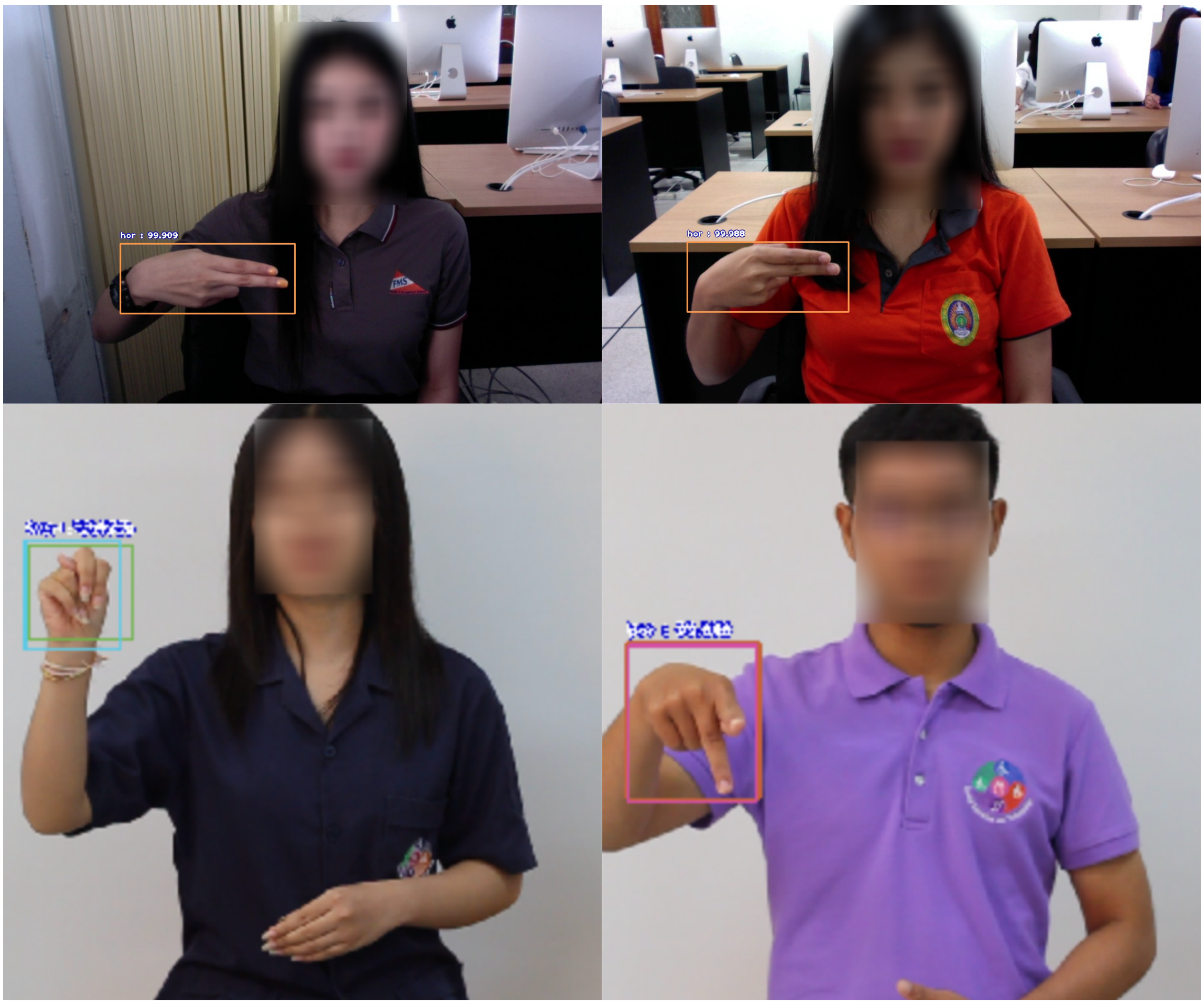

**Fig. 9.** Examples of hand detection images using the YOLOv3 algorithm.

# LIMITATIONS

Not applicable

# ETHICS STATEMENT

These images were collected with the objective of recognizing Thai one-stage fingerspelling. The study received approval from the local ethics committee at Mahasarakham University (Document No. 360-355/2564).

# CRediT AUTHOR STATEMENT

**Siriwiwat Lata:** Conceptualization, Data Curation, Investigation, Methodology, Resources, Validation, Writing – Original Draft; **Sirawan Phiphitphatphaisit:** Data Curation, Methodology, Validation; **Emmanuel Okafor:** Conceptualization, Validation, Writing – Review & Editing; **Olarik Surinta:** Supervision, Conceptualization, Methodology, Experimental Design, Writing – Review & Editing, Funding Acquisition.

## ACKNOWLEDGEMENTS

This research project was financially supported by Mahasarakham University, Thailand.

## DECLARATION OF COMPETING INTERESTS

The authors declare that they have no known competing financial interests or personal relationships that could have appeared to influence the work reported in this paper.